\documentclass{article}

\usepackage[preprint]{neurips_2025}

\usepackage[dvipsnames]{xcolor}

\usepackage{listings}

\lstdefinestyle{mypython}{
  language=Python,
  basicstyle=\ttfamily\footnotesize,
  keywordstyle=\color{blue},
  commentstyle=\color{gray},
  stringstyle=\color{orange},
  numbers=left,
  numberstyle=\tiny\color{gray},
  stepnumber=1,
  numbersep=6pt,
  showstringspaces=false,
  breaklines=true,
  tabsize=4
}

\usepackage[utf8]{inputenc} % allow utf-8 input
\usepackage[T1]{fontenc}    % use 8-bit T1 fonts
\usepackage{url}            % simple URL typesetting
\usepackage{booktabs}       % professional-quality tables
\usepackage{amsfonts}       % blackboard math symbols
\usepackage{nicefrac}       % compact symbols for 1/2, etc.
\usepackage{microtype}      % microtypography
\usepackage{xcolor}         % colors
\usepackage{graphicx}
\usepackage{amsmath} 
\usepackage{xspace} 
\usepackage{wrapfig}
\usepackage{multirow}
\usepackage[parfill]{parskip}

\usepackage{tikz}
\usepackage{amsmath}
\usepackage{amssymb}
\usepackage{pifont}
\usepackage{xcolor}
\usepackage{colortbl}
\usepackage{booktabs}
\usepackage{multirow}
\usepackage{mathrsfs}
\usepackage{marvosym}
\usepackage{enumerate}
\usepackage[hang,flushmargin,para]{footmisc} 
\usepackage{tablefootnote}

\usepackage{booktabs}
\usepackage{multirow}
\usepackage{makecell}
\usepackage{subcaption}
\usepackage{array}
\usepackage{graphicx}
\usepackage{bm}
\usepackage{bbm}

\definecolor{darkgreen}{RGB}{50,200,0}
\definecolor{darkred}{RGB}{200, 0, 0}
\definecolor{Gray}{gray}{0.93}
\definecolor{uclagold}{rgb}{1.0, 0.7, 0.0}
\definecolor{airforceblue}{rgb}{0.36, 0.54, 0.66}
\definecolor{rosegold}{rgb}{0.72, 0.43, 0.47}
\definecolor{pastelbrown}{rgb}{0.51, 0.41, 0.33}
\definecolor{isabelline}{rgb}{0.96, 0.94, 0.93}
\definecolor{macaroniandcheese}{rgb}{0.98, 0.89, 0.83}
\definecolor{wildblueyonder}{rgb}{0.85, 0.89, 0.95}
\definecolor{mediumtaupe}{rgb}{0.4, 0.3, 0.28}
\definecolor{bluegray}{rgb}{0.4, 0.6, 0.8}
\definecolor{celestialblue}{rgb}{0.29, 0.59, 0.82}
\definecolor{darkorange}{rgb}{1.0, 0.55, 0.0}
\definecolor{cadmiumred}{rgb}{0.89, 0.0, 0.13}
\definecolor{magnolia}{rgb}{0.97, 0.96, 1.0}
\definecolor{pastelblue}{rgb}{0.68, 0.78, 0.81}
\definecolor{persiangreen}{rgb}{0.0, 0.65, 0.58}
\definecolor{steelblue}{rgb}{0.27, 0.51, 0.71}
\definecolor{bluebell}{rgb}{0.64, 0.64, 0.82}
\definecolor{dimgray}{rgb}{0.41, 0.41, 0.41}
\definecolor{splashedwhite}{rgb}{1.0, 0.99, 1.0}
\definecolor{lavendergray}{rgb}{0.77, 0.76, 0.82}
\definecolor{lightgray}{rgb}{0.92, 0.92, 0.92}
\definecolor{lavendermist}{rgb}{0.9, 0.9, 0.98}
\definecolor{lightgreen}{HTML}{f8fcf4}
\definecolor{lightblue}{HTML}{dfebf7}
\definecolor{zeroshot}{rgb}{0.9, 0.9, 0.9}
\definecolor{fourshot}{rgb}{0.8, 0.9, 0.8}
\definecolor{eightshot}{rgb}{0.8, 0.8, 0.9}
\definecolor{sixteenshot}{rgb}{0.9, 0.8, 0.8}

\usepackage{cite}
\definecolor{citeblue}{HTML}{0071bc}
\definecolor{textpurple}{RGB}{135,89,201}

\definecolor{mycitecolor}{rgb}{0, 0.4, 0.7}
\usepackage[hidelinks,colorlinks=true,citecolor=mycitecolor]{hyperref}

\newcommand{\model}{LaVi\xspace}
\newcommand{\modelbase}{LaVi-Image\xspace}
\newcommand{\modelhighres}{LaVi-Image (HD)\xspace}
\newcommand{\modelvideo}{LaVi\xspace}

\title{\model: Efficient Large Vision-Language Models via Internal Feature Modulation}

\author{Tongtian Yue\textsuperscript{1,2\thanks{Equal Contribution.}}, \quad Longteng Guo\textsuperscript{1,2*}, \quad Yepeng Tang\textsuperscript{3*}, \quad \\ \textbf{Zijia Zhao}\textsuperscript{1,2}, \quad \textbf{Xinxin Zhu}\textsuperscript{1,2}, \quad \textbf{Hua Huang}\textsuperscript{4}, \quad \textbf{Jing Liu}\textsuperscript{1,2\thanks{Corresponding author.}} \\
\textsuperscript{1}Institute of Automation, Chinese Academy of Sciences \\
\textsuperscript{2}School of Artificial Intelligence, University of Chinese Academy of Sciences
\\ \textsuperscript{3}Institute of Information Science, Beijing Jiaotong University
\\ \textsuperscript{4}School of Artificial Intelligence, Beijing Normal University}

\begin{document}
\maketitle
\begin{abstract}
Despite the impressive advancements of Large Vision-Language Models (LVLMs), existing approaches suffer from a fundamental bottleneck: inefficient visual-language integration. Current methods either disrupt the model's inherent structure or introduce severe long-context computational burden, severely limiting scalability and efficiency. In this paper, we rethink multimodal integration and present \model, a novel LVLM that enables seamless and efficient vision-language fusion through internal feature modulation within the Large Language Models (LLMs). Unlike dominant LVLMs that rely on visual token concatenation, \model bypasses long-context expansion by introducing a lightweight and adaptive transformation,  which incorporates visual context by injecting token-wise vision-conditioned \textit{deltas} into the affine parameters of layer normalization. This mechanism directly modulates linguistic hidden states based on visual input, ensuring precise vision-language alignment while preserving the LLM’s linguistic priors and drastically reducing computational costs. Extensive evaluations across 15 image and video benchmarks demonstrate that \model not only achieves state-of-the-art multimodal performance but also dramatically enhances efficiency. Compared to LLaVA-OV-7B, \model reduces FLOPs by 94.0\%, improves inference speed by 3.1×, and cuts memory usage in half — establishing \model as a scalable and practical solution for real-time multimodal reasoning. The code and models will be released soon.

% Large Vision-Language Models (LVLMs) have demonstrated strong capabilities across image and video understanding tasks, yet their practical deployment remains hindered by inefficient visual-language integration. Existing strategies either compromise the pretrained structure of Large Language Models (LLMs) through intrusive architectural modifications or suffer from substantial computational overhead due to the concatenation of large visual token sequences into the context. In this work, we introduce \model, an efficient LVLM that addresses the challenge of vision-language fusion via Feature Modulation Injection (FMI) — a lightweight integration paradigm that modulates the internal hidden states of the LLM based on visual input. Central to our design is Vision-Infused Layer Normalization (ViLN), which incorporates visual context by injecting token-wise vision-conditioned \textit{deltas} into the affine parameters of layer normalization. This approach maintains the architectural integrity of LLMs, preserves linguistic priors, and circumvents the quadratic complexity of visual token attention. \model supports flexible visual conditioning and achieves fine-grained multimodal alignment with minimal overhead. It surpasses or matches state-of-the-art performance across 15 benchmarks while reducing FLOPs by 94.0\%, improving inference speed by 3.1×, and cutting memory usage in half compared to LLaVA-OV-7B — establishing \model as a scalable and practical solution for real-time multimodal reasoning.

\end{abstract}    
\section{Introduction}
\label{sec:intro}

Recently, significant advancements in Large Language Models (LLMs)~\citep{radford2019language, achiam2023gpt, yang2024qwen2, touvron2023llama} have catalyzed the emergence of Large Vision-Language Models (LVLMs)~\citep{bai2023qwen, liu2023improvedllava, awadalla2023openflamingo, tong2024cambrian}, demonstrating remarkable capabilities in visual perception and cognitive reasoning~\citep{liu2024mmbench, fu2023mme, singh2019towards, hudson2019gqa}. While considerable progress has been achieved separately in visual encoding and language generation, the pivotal challenge of effectively integrating visual information into LLMs still remains open. 

Existing integration techniques generally fall into two categories. The first, termed  \textbf{\textit{architectural injection}} (\textit{e.g.}, Flamingo~\citep{awadalla2023openflamingo}), augments the original LLMs by introducing additional layers~\citep{Alayrac2022FlamingoAV, meta2024llama, ye2024mplug}, such as cross-attention and feed-forward layers, strategically throughout the model. 
While these modules explicitly insert visual features into the linguistic processing pathway, their introduction inherently disrupts the architectural coherence and processing flow of the original LLMs. Consequently, it can degrade the delicate pre-trained language understanding, risking losing the rich linguistic priors encoded within LLMs~\citep{zhang_wings,luo2024mono,wang2025cogvlm}. 
The second and currently predominant approach, \textbf{\textit{in-context injection}} (\textit{e.g.}, the LLaVA series~\citep{liu2024llavanext, liu2023improvedllava, li2024llava}), integrates visual information by concatenating vision-derived token sequences directly into textual input, treating them as part of the initial context for the LLMs. While preserving architectural integrity, this method introduces significant practical challenges. Specifically, the large number of visual tokens required (\textit{e.g.}, 576 tokens for a single image using CLIP ViT-L/336px~\citep{radford2021learning}) leads to severe computational overhead due to the quadratic complexity inherent in self-attention mechanisms~\citep{vaswani2017attention}. This complexity escalates dramatically when processing high-resolution images or long video sequences, resulting in substantial inference latency and computational bottlenecks, thus hindering real-time applicability.

Through analyzing these methods, we argue that an ideal visual-language integration strategy must satisfy two fundamental principles:
1) \textbf{\textit{minimal structural interference}}, which ensures the preservation of pretrained linguistic knowledge to support coherent text generation and empower vision-grounded understanding and reasoning; and 2) \textbf{\textit{computational scalability}}, which mitigates inefficiencies arising from quadratic complexity when processing extensive visual tokens.

Guided by these principles, we propose a new visual-language integration strategy for LVLMs: internal \textbf{\textit{Feature Modulation Injection}} (FMI) within the LLMs. Central to FMI is the layer normalization (LN) mechanism~\citep{ba2016layer, zhang2019root}, a ubiquitous component in modern LLMs. LN applies learnable affine transformations to rescale and shift the features, offering a natural pathway for updating the internal hidden states through a combination of additive and multiplicative modulation. Inspired by this, we introduce Vision-Infused Layer Normalization (ViLN), a lightweight yet effective extension of standard LN that integrates visual context into language modeling. Specifically, ViLN dynamically adjusts the affine transformation parameters of LN via vision-conditioned \textit{deltas}, thereby enabling visual signals to modulate token-wise linguistic representations in a fine-grained and context-aware manner. Based on ViLN, FMI provides notable advantages by minimally intervening within the pretrained LLM, thus preserving its linguistic priors and relieving the impact on linguistic performance. Moreover, by avoiding direct incorporation of visual tokens into self-attention, our method circumvents the quadratic complexity issue, achieving superior computational scalability and efficiently accommodating extensive visual data such as high-resolution images and long videos.

Building upon this strategy, we present \textbf{\model} (Language and Vision Integrator), a novel LVLM that seamlessly incorporates FMI through selective replacement of standard LN layers with ViLN modules. To supply ViLN with the necessary visual conditions, \model employs a conditioning module that generates a dedicated visual condition for each text token, allowing token-wise modulation informed by the visual features. The design of this conditioning module is highly flexible, and we explore three alternative implementations: MLP-based, convolution-based, and attention-based approaches, each offering a favorable trade-off between computational efficiency and multimodal performance. Once the visual conditions are obtained, each is transformed into vision \textit{deltas} of the affine parameters through a lightweight learnable mapping, which are then used to modulate the corresponding linguistic features within the LLM.

\begin{wrapfigure}{rt!}{0.585\textwidth}
    \centering
    \vspace{-\baselineskip}\setlength\intextsep{0pt}
    \includegraphics[width=\linewidth]{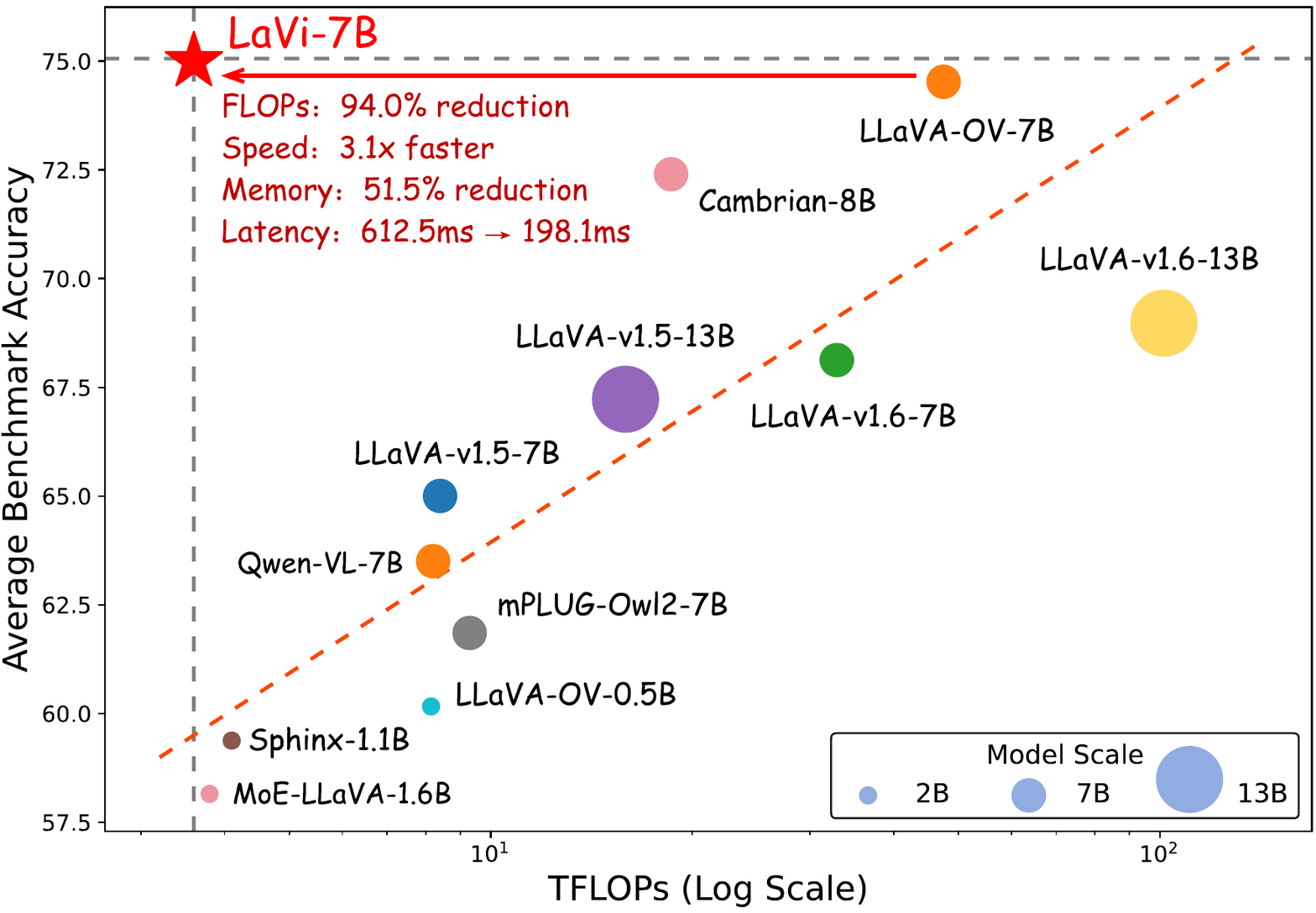}
  \caption{\textbf{Comparison between \model and open-source LVLMs on image understanding benchmarks.} We report the average accuracy on MMBench~\citep{liu2024mmbench}, MME~\citep{fu2023mme}, TextVQA~\citep{singh2019towards}, and GQA~\citep{hudson2019gqa}. For MME, scores are normalized to percentages. The red dashed line represents the linear fit to all models except \model.}
  \label{fig:intro}
  \vspace{-10pt}
\end{wrapfigure}

Benefiting from a significantly reduced context length and a lightweight yet effective visual-language integration strategy, \model strikes an impressive balance between computational efficiency and benchmark performance. Comprehensive evaluations across 9 image-based and 6 video-based understanding benchmarks demonstrate that \model achieves state-of-the-art performance comparable to LLaVA-style models while substantially reducing computational overhead. Moreover, it maintains superior linguistic capabilities compared to using other injection strategies. Compared to the baseline LLaVA-OV-7B~\citep{li2024llava}, \model, despite maintaining the same 7B parameter scale, demonstrates substantial improvements in both efficiency and performance. It achieves an impressive 94.0\% reduction in FLOPs, operates 3.1× faster, lowers memory consumption by 51.5\%, and reduces inference latency from 612.5 ms to just 198.1 ms, as illustrated in Figure \ref{fig:intro}. Remarkably, \model requires even fewer FLOPs than LLaVA-OV-0.5B~\citep{li2024llava}, yet surpasses it by +15.5 points in benchmark accuracy. These advancements significantly enhance real-time multimodal interactions, positioning \model as a highly efficient alternative in the evolving landscape of LVLMs.

Our contribution can be concluded as:

\begin{itemize}
\item We introduce a novel internal feature modulation injection paradigm for LVLMs. It ensures minimal structural interference, effectively preserves pretrained linguistic priors, and achieves computational scalability by avoiding excessive context length expansion.

\item We propose \model, a highly efficient LVLM capable of comprehensive image and video understanding. \model integrates ViLN to achieve fine-grained visual-linguistic alignment and investigates various visual conditioning mechanisms, effectively balancing multimodal performance with computational efficiency.

\item \model achieves state-of-the-art performance across multimodal benchmarks while significantly improving computational efficiency. Compared to the LLaVA-style baseline, \model reduces FLOPs by 94.0\%, offering an efficient and practical solution for real-time multimodal processing with significantly reduced resource demands.
\end{itemize}

% To this end, we propose \model, 一个高效且简单的LVLMs架构。它完整保留了LLM原始的处理流程。具体而言，它仅将文本序列输入LLM，保持LLM原始处理流程的同时避免了任何特定交互模块的引入。\model实现视觉conditioning的方式是简单但有效的。Technically, 它通过Layer Norm这一常见于所有LLMs的模块来注入视觉信息。具体而言，Layer Norm中存在用以仿射变换的可学习参数，用来对归一化后的文本特征进行shift或者scale操作。\model的操作仅仅是让这些参数的学习变得condition化。即使用视觉特征regress得到这些参数。考虑到这些参数本身的轻量性以及其与文本特征的交互仅仅是稀疏且简易的加性和乘性操作，因此\model实现跨模态交互的方式非常高效。此外，\model对于LLM原始架构和流程的改动是minimal的，本质上只是改变了LLM内部少量参数的可学习方式。
% 实验结论

\section{Related Work}

\textbf{Large Vision-Language Models.}
% Recent advancements in LVLMs have significantly reshaped multimodal understanding by seamlessly integrating visual perception with linguistic reasoning. Prominent commercial models, including GPT-4V~\citep{achiam2023gpt}, Claude 3.5~\citep{anthropic2024claude}, and Gemini~\citep{team2023gemini}. Besides, there are some open-source initiatives such as LLaVA~\citep{liu2023improvedllava, liu2024llavanext, li2024llava}, Qwen-VL~\citep{bai2023qwen,yang2024qwen2}, InternVL~\citep{chen2024internvl, chen2024far}, and BLIP series~\citep{li2022blip, li2023blip}. These foundational capabilities in vision-language processing are progressively extending to domains involving high-resolution images~\citep{liu2024sphinx, guo2024llava, zhang2024beyond} and long-duration videos~\citep{zhang2023videollama, video-chat, maaz2023video}. However, this approach is inherently constrained by the quadratic complexity of self-attention mechanisms~\citep{vaswani2017attention}, leading to excessive memory consumption and inference latency, especially when processing high-resolution images or long video sequences. In contrast, we integrate visual information through internal feature modulation, where the LLM's hidden states are adaptively modulated without expanding context length. This approach avoids quadratic complexity, ensuring efficient and scalable multimodal processing.

Large Vision-Language Models (LVLMs) have significantly advanced multimodal understanding, enabling more effective integration of vision and language. Closed-source models such as Claude~\citep{anthropic2024claude}, GPT~\citep{achiam2023gpt}, and Gemini~\citep{team2023gemini} series exhibit strong multimodal capabilities. Meanwhile, open-source models like LLaVA~\citep{llava, liu2023improvedllava, liu2024llavanext, li2024llava}, BLIP~\citep{li2022blip, li2023blip}, Qwen-VL~\citep{bai2023qwen,yang2024qwen2}, and InternVL~\citep{chen2024internvl, chen2024far} series have contributed significantly to the community by providing accessible and adaptable alternatives. Recent research has focused on improving input resolution~\citep{liu2024llavanext, guo2024llava}, enhancing training and inference efficiency~\citep{fastv, wan2024look}, and extending multimodal capabilities to temporal video sequences~\citep{video-chat, zhang2023videollama, maaz2023video}. These advancements continue to push the boundaries of LVLMs in efficiency, scalability, and multimodal reasoning.

\textbf{Multimodal Integration.} 
Multimodal integration, which aims to effectively fuse visual and language representations, is a fundamental challenge in multimodal learning. A common approach is to directly concatenate or perform element-wise operations (\textit{e.g.}, addition or multiplication) on vision and language features, which has been widely adopted in tasks such as visual question answering (VQA)~\citep{antol2015vqa, jabri2016revisiting} and vision-language representation learning~\citep{chen2020uniter, lu2019vilbert}. In the LVLM domain, LLaVA~\citep{llava} also employs this straightforward method, inspiring many subsequent works. While this approach is computationally simple, it significantly increases sequence length, posing efficiency challenges. An alternative strategy involves using additional fusion structures, such as cross-attention mechanisms~\citep{anderson2018butd, yu2022coca} or multi-layer transformers~\citep{li2022blip, albef}, which introduce learnable parameters to compress multimodal representations more effectively. However, such modifications (\textit{e.g.}, Flamingo~\citep{awadalla2023openflamingo}) can disrupt the pretrained data flow of the underlying LLM, potentially degrading language capabilities or increasing complexity.
\section{Methodology}

\subsection{Preliminaries}
\begin{figure*}[t]
    \centering
    \includegraphics[width=\linewidth]{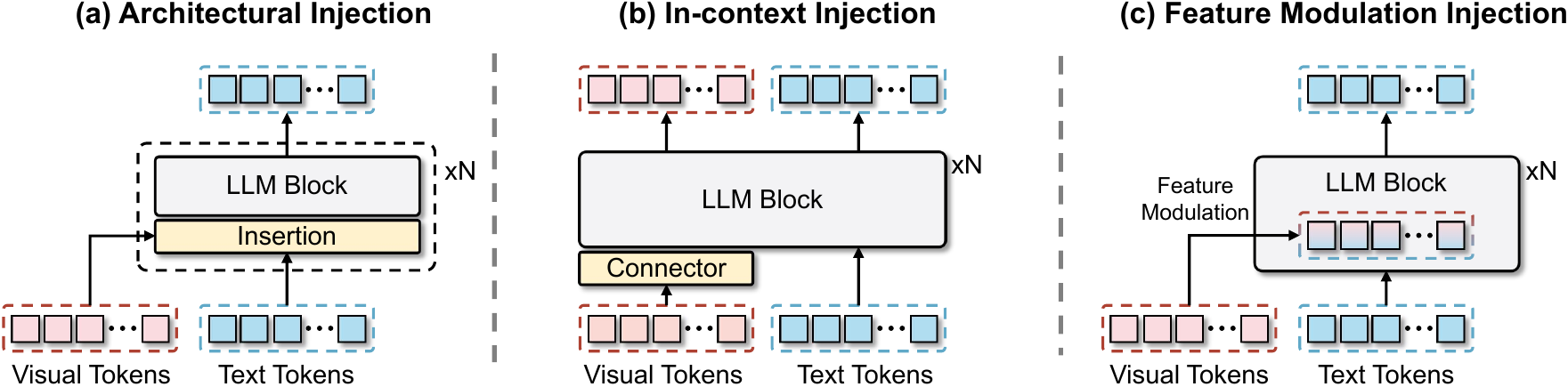} % 占位图
    \vspace{-10pt}
    \caption{\textbf{Comparisons of various vision integration techniques for LVLMs.} (a) Architectural injection: additional layers are inserted into LLM for cross-modal interaction;  (b) In-context injection: visual tokens are concatenated before the text sequence as the initial context. (c) Feature modulation injection (Ours): the internal hidden states are 
    modulated by the vision-guided affine transformation.}
    \label{fig:pre}
    \vspace{-10pt}
\end{figure*}

In this section, we begin with a concise overview of the predominant visual-language integration strategies employed in LVLMs. Specifically, existing methods primarily fall into two categories: architectural injection and in-context injection: 

\textbf{Architectural Injection.} As illustrated in Figure \ref{fig:pre}\textcolor{red}{a}, this approach integrates visual information by inserting additional interaction layers (\textit{e.g.}, cross attention~\citep{awadalla2023openflamingo, Alayrac2022FlamingoAV} and hyper attention ~\citep{ye2024mplug}), enabling fusion between the text sequence $\bm{t}$ and visual features $\bm{v}$ within the $\Theta$-parameterized LLM:
\begin{equation}
    \mathbf{H}_0 = \bm{t}, \quad\mathbf{H}_{\ell+1} = \Theta_{\ell} \big( \Phi_{\ell}(\bm{v}, \mathbf{H}_\ell) \big)
\end{equation}
where $\mathbf{H}_{\ell}$ denotes the hidden states at layer $\ell$, $\Phi_{\ell}(\cdot, \cdot)$ represents the inserted cross-modal interaction module, and $\Theta_{\ell}(\cdot)$ denotes the $\ell$-th layer of the LLM. While this method ensures direct multimodal alignment, it comes at the cost of architectural disruption, requiring extensive modifications to the pretrained LLMs. Such modifications compromise the model’s linguistic priors, potentially degrading the generative capabilities in both multimodal and language-only contexts.

\textbf{In-context Injection.} 
As illustrated in Figure \ref{fig:pre}\textcolor{red}{b}, this approach involves mapping visual features $\bm{v}$ into the LLM’s semantic space via a vision-language connector~\citep{li2023blip, li2024llava, liu2024llavanext, liu2023improvedllava} and appending them as a visual prefix before the text sequence $\bm{t}$:
\begin{equation}
    \mathbf{H}_0 = [\bm{v}; \bm{t}], \quad\mathbf{H}_{\ell+1} = \Theta_{\ell} \big( \mathbf{H}_\ell \big)
\end{equation}
This method allows cross-modal interaction to occur within the LLM’s existing self-attention layers, avoiding explicit structural modifications. However, because self-attention scales quadratically with sequence length~\citep{vaswani2017attention}, the introduction of numerous visual tokens leads to severe computational inefficiencies. This becomes particularly problematic when processing high-resolution images or long video sequences, where the number of visual tokens grows significantly.

To address the inefficiencies of these approaches, we propose feature modulation injection (FMI), as depicted in Figure \ref{fig:pre}\textcolor{red}{c}. Instead of injecting additional layers or expanding sequence length, FMI incorporates visual information directly into the internal hidden states of the LLM via a lightweight modulation mechanism. More details are provided in the following section.

% Specifically, FMI replaces standard Layer Normalization with a vision-conditioned transformation module. Given extracted visual features, a token-wise conditioning function generates scaling ($\alpha$) and shifting ($\beta$) parameters for each token, which dynamically adjust the hidden states of the LLM. This mechanism ensures precise vision-language alignment while preserving the original LLM architecture. By eliminating the need for lengthy visual token sequences, it significantly enhances computational efficiency while maintaining the linguistic priors.

% Vision-Infused Layer Norm (ViLN), which dynamically rescales and shifts linguistic hidden states based on vision-conditioned transformations.

% We provide the details in the following section.
\subsection{Feature Modulation Injection.}
 At the core of FMI is the Layer Normalization (LN) module~\citep{ba2016layer, zhang2019root}, a ubiquitous and essential component in virtually all mainstream LLM architectures. Given an input text sequence $\bm{t} = \{\bm{t}_i\}_{i=1}^T$, a typical LLM block processes $\bm{t}$ as follows:
\begin{equation}
\bm{t} \leftarrow \bm{t} + \mathcal{F}_{\text{\textit{att}}}(\text{LN}_1(\bm{t}))
\label{eq:ln_attn}
\end{equation}
\begin{equation}
\bm{t} \leftarrow \bm{t} + \mathcal{F}_{\text{\textit{ffn}}}(\text{LN}_2(\bm{t}))
\label{eq:ln_ffn}
\end{equation}
Here, $\mathcal{F}_{\text{\textit{att}}}$ and $\mathcal{F}_{\text{\textit{ffn}}}$ denote the self-attention and feed-forward sub-layers, respectively. The LN module normalizes the input features via:
\begin{equation}
\text{LN}(\bm{t}) = \alpha \odot \frac{\bm{t} - \mu}{\sigma} + \beta = \alpha \odot \hat{\bm{t}} + \beta
\end{equation}
where $\mu$ and $\sigma$ are the mean and standard deviation of $\bm{t}$, and $\alpha$, $\beta$ are learnable affine parameters that control the scaling and shifting of the normalized features. Inspired by this, we propose to \textit{link the learning of affine parameters to visual features}, thereby making the modulation of internal text representations within LN explicitly vision-conditioned. Specifically, we define the following Vision-Infused Layer Normalization (ViLN):
\begin{equation}
\label{eq:vln}
\text{ViLN}(\bm{t}, \bm{v}) = (\alpha + \Delta \alpha_{\bm{v}}) \odot \hat{\bm{t}} + (\beta + \Delta \beta_{\bm{v}}),
\end{equation}
Here, $\Delta \alpha_{\bm{v}}$ and $\Delta \beta_{\bm{v}}$ are vision-guided modulation parameters that adaptively adjust the original affine parameters $\alpha$ and $\beta$ in LLM based on visual context. They are dynamically regressed from visual features $\bm{v}$ through a token-wise conditioning module, which will be detailed later. 

Overall, FMI transforms visual information into affine parameters that directly adjust the LLM’s internal hidden states via \textit{multiplicative} and \textit{additive} operations. This enables a direct and efficient fusion of vision information at the feature level, eliminating the need for lengthy visual token sequences or additional cross-modal interaction modules.

\subsection{LaVi: A Highly Efficient LVLM}
\label{sec:model}

 The overall architecture of \model is illustrated in Figure~\ref{fig:model}. After replacing the internal LN of the LLM with ViLN, \model leverages a conditioning module to generate token-wise affine parameters from visual features $\bm{v}$, comprising two sets that are applied before the self-attention and feed-forward sublayers, respectively:
\begin{equation}
[\Delta \alpha_{\bm{v}}^1, \Delta \beta_{\bm{v}}^1, \Delta \alpha_{\bm{v}}^2, \Delta \beta_{\bm{v}}^2] = \text{Swi}\big( \text{Cond}(\bm{t}, \bm{v}) \big) \mathbf{W} + \mathbf{b}
\end{equation}
Here, $\text{Swi}(\cdot)$ denotes the Swish activation function~\citep{ramachandran2017searching}, while $\mathbf{W}$ and $\mathbf{b}$ are learnable projection weights and bias, respectively. This projection is zero-initialized to ensure that the vision-conditioned \textit{deltas} are initially zero, so that the forward pass exactly replicates the original LLM behavior—thereby facilitating stable adaptation and linguistic priors preservation during early training. 

The conditioning function $\text{Cond}(\cdot)$ is responsible for aggregating visual context relevant to each token in the text sequence $\bm{t}$. The design of this function is highly flexible. In this paper, we explore three alternative instantiations:

\begin{figure*}[t]
    \centering
    \includegraphics[width=0.9\textwidth]{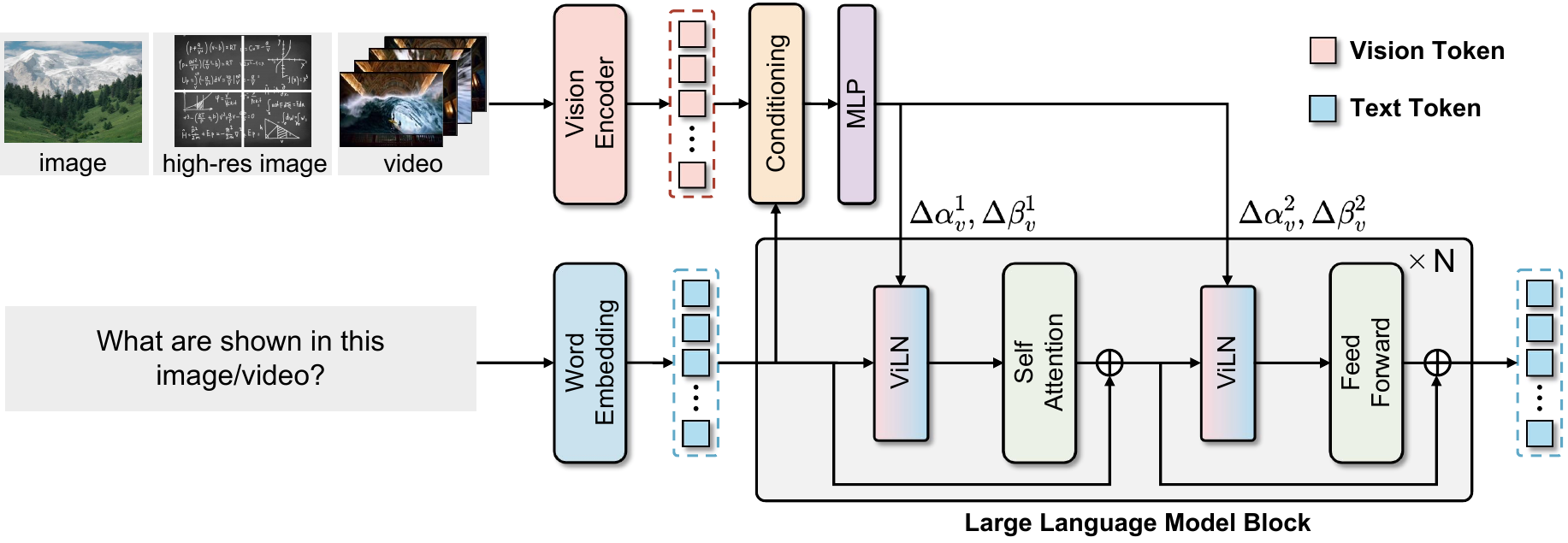} % 占位图
    \caption{\textbf{An illustrative diagram of the overall model architecture.} For a LLM block equipped with ViLN, visual and textual features are fed into the conditioning module to obtain token-wise visual conditions. Through a lightweight MLP, these conditions are then transformed into scale and shift parameters, which modulate the internal linguistic features of the LLM.}
    \label{fig:model}
    \vspace{-10pt}
\end{figure*}

\textbf{MLP-based Conditioning}. Inspired by MLP-Mixer~\citep{tolstikhin2021mlp}, we design two sequential MLPs to aggregate visual context. Given a text token $t_i$, we concatenate it with visual features $\bm{v}$, then transpose the sequence $[t_i;\bm{v}]$ to interchange token and channel dimensions. A token-mixing MLP integrates information across tokens, and after transposing back, a channel-mixing MLP blends features across dimensions. The vision-aware embedding for $t_i$ is then extracted at original position:
\begin{equation}
\text{Cond}_{mlp}(t_i,\bm{v})
\;=\;
\Bigl[
\mathbf{MLP}_{channel}
\bigl(
\bigl(\mathbf{MLP}_{token}([\;t_i;\bm{v}\,]^{\top})\bigr)^{\top}
\bigr)
\Bigr]_{t_i}
\end{equation}
\textbf{Conv-based Conditioning}. Inspired by ConvMixer~\citep{trockman2022patches}, we treat the concatenated sequence $[t_i; \bm{v}]$ as a 1-D signal along the token dimension. We first apply a depth-wise convolution followed by an activation $\sigma$ to mix information between $t_i$ and visual features $\bm{v}$. Subsequently, a point-wise convolution is utilized to integrate these features across the embedding dimension. The resulting representation at the token position corresponding to $t_i$ provides the vision-aware embedding:
\begin{equation}
{\;
\text{Cond}_{conv}(t_i,\bm{v})
=\Bigl[
\mathbf{Conv}_{point}\!\bigl(\,
\sigma\!\bigl(
\mathbf{Conv}_{depth}\bigl([\;t_i;\bm{v}\,]\bigr)
\bigr)\bigr)
\Bigr]_{t_i}
\;}
\end{equation}
\textbf{Attention-based Conditioning.} We introduce a cross-attention module, where text token $t_i$ is used as the query, while visual tokens $\bm{v}$ serve as keys and values. Through the attention mechanism, we directly aggregate relevant visual context to produce the vision-aware representation for $t_i$:
\begin{equation}
\text{Cond}_{attn}(t_i, \bm{v}) = \text{Attention}(t_i \mathbf{W}_Q, \bm{v} \mathbf{W}_K, \bm{v} \mathbf{W}_V)
\end{equation}
We provide further implementation details for the three paradigms in the Appendix. By default, we adopt the attention-based approach due to its simplicity and effectiveness. In Section~\ref{sec:arc_compare}, we provide a comparative analysis of each conditioning function, demonstrating that all approaches provide robust multimodal integration with minimal computational overhead.

\textbf{Scalable Visual Input Support.}
\model flexibly accommodates more complex visual inputs—such as high-resolution images and videos—while requiring only minimal structural modifications, making it broadly applicable across diverse vision-language scenarios. Specifically, for high-resolution images, we adopt a tiling strategy, where the image is divided into non-overlapping tiles compatible with the native input size of the vision encoder. Each tile is independently encoded, and the resulting visual tokens are concatenated along the sequence dimension. For videos, we uniformly sample $k$ frames. Each frame is encoded by the vision encoder and undergoes 2$\times$2 adaptive pooling. The resulting frame features are concatenated sequentially, with shared temporal position encoding applied to each frame's tokens to capture temporal dynamics.

% \subsection{High-Res Image and Video Extension}
% \label{sec:extension}
% \model can be seamlessly extended to high-resolution images and videos with minimal structural modifications.

% \textbf{High-Resolution Image.} For high-resolution images, we follow the method in LLaVA-v1.6~\citep{liu2024llavanext}, where the image is divided into smaller grids matching the vision encoder's trained resolution. Each grid is independently encoded, and the resulting feature maps are concatenated along the sequence dimension to form a unified feature map.

% \textbf{Video.} For video processing, we follow the standard setup used in current video-language models~\citep{lin2023video,chen2024sharegpt4video, damonlpsg2024videollama2, li2024mvbench}, uniformly sampling $k$ frames from the video. Each frame is encoded by the vision encoder and undergoes 2 adaptive pooling. The resulting frame features are concatenated sequentially, with shared temporal position encoding applied to each frame's tokens to capture temporal dynamics. Unlike dedicated video-language models, we do not introduce video-specific architectural modifications, as our primary focus is on efficient vision-language integration. Nevertheless, this straightforward approach demonstrates strong effectiveness, achieving performance on par with SoTA models.

\section{Experiments}

\begin{table*}[t]
    \centering
    \caption{\textbf{Performance on 9 image-based benchmarks,} including VQAv2, GQA, VisWiz, ScienceQA, TextVQA, POPE, MME$^\text{P}$, MMBench and SEED$^\text{I}$. For MME$^\text{P}$, the scores are presented as percentages. Along with efficiency and accuracy, we also report the LLM backbone for each baseline. } 
    \setlength{\tabcolsep}{2.6pt} % 缩小列间距
    \resizebox{\linewidth}{!}{  
    \begin{tabular}{ll |cc| ccccccccccc}
        \toprule
        \multirow{2}{*}{\textbf{Method}} & \multirow{2}{*}{\textbf{LLM}} &
        \multicolumn{2}{c|} {\textbf{Efficiency}} & \multicolumn{10}{c} {\textbf{Performance}} \\ 
        \cmidrule(lr){3-4} \cmidrule(lr){5-14}

        & & \textbf{FLOPs} & \textbf{Latency} & \textbf{VQA$^\text{v2}$} & \textbf{GQA} & \textbf{VisWiz} & \textbf{SciQA} & \textbf{VQA$^\text{T}$} & \textbf{POPE} & \textbf{MME$^\text{P}$} & \textbf{MMB} & \textbf{SEED$^\text{I}$}  & \textbf{Avg.} \\
        \midrule 
        % \rowcolor{gray!15}
        \multicolumn{13}{l}{\textbf{\textit{Baselines with $\leq$ 2B parameters scale}}} \\
        MoE-LLaVA \citep{lin2024moe} & StableLM-1.6B &  3.8 & 206.4 & 76.0 &  60.4 & 37.2 & 62.6 & 47.8 &  84.3 & 65.0 & 59.4 & --  & -- \\
        MobileVLM-V2 \citep{chu2024mobilevlm} & MLLaMA-1.4B & 4.3 & 214.9 & -- &  59.3 & -- & 66.7 & 52.1&  84.3 & 65.1 & 57.7 & -- & -- \\
        SPHINX-tiny \citep{liu2024sphinx} & TLLaMA-1.1B & 4.1 & 212.3 & 74.7 & 58.0 & 49.2 & 21.5 & 57.8 & 82.2 & 63.1 & 56.6 & 25.2 & 54.3 \\
        LLaVA-OV \citep{li2024llava} & Qwen2-0.5B & 7.8 &  228.0 & 78.5 & 58.0 & 51.4 & 67.2 & 65.9 & 86.0& 61.9 & 52.1 & 65.5 & 65.2 \\
        \midrule 
        % \rowcolor{gray!15}
        \multicolumn{13}{l}{\textbf{\textit{Baselines with $\leq$ 8B parameters scale}}} \\
        % Flamingo \citep{awadalla2023openflamingo} &  MPT-7B & -- & -- & 53.0 & -- & -- & 44.8 & 28.3 & -- & -- & 12.7 & 42.7 & --\\
        % mPLUG-Owl3 \citep{ye2024mplug} & Qwen2-7B & -- & -- & 82.5 & 64.6 & -- & -- & 69.0 & 88.2 & -- & 77.6 & -- & -- \\
        Qwen-VL-Chat \citep{bai2023qwen} & Qwen-7B & 8.2 & 239.4 & 78.2 & 57.5 & 38.9 & 68.2 & 61.5 & –  & 74.4 & 60.6 &  65.4 & -- \\
        mPLUG-Owl2 \citep{ye2023mplug} &  LLaMA2-7B & 9.3 & 278.6 & 79.4 & 56.1 & 54.5 & 68.7 & 54.3 & -- & 72.5 & 64.5 & 57.8 & -- \\
        Cambrian-1 \citep{tong2024cambrian} & LLaMA3-8B & 18.6 & 393.7 & -- & 64.6 & -- & 80.4 & 71.7 & -- & 77.4 & 75.9 & 74.7 & -- \\
        LLaVA-v1.5 \citep{liu2023improvedllava} & Vicuna-7B & 8.4 &  254.4 & 78.5 & 62.0 & 50.0 & 66.8 & 58.2 & 85.9 & 75.5 & 64.3 & 66.1 & 67.5 \\
        LLaVA-v1.6 \citep{liu2024llavanext} & Vicuna-7B & 32.9 & 502.4 & 81.8 & 64.2 & 57.6 & 70.1 & 64.9 & 86.5 & 76.0 & 67.4 & 70.2 & 71.0 \\
        LLaVA-OV \citep{li2024llava} & Qwen2-7B & 60.4 & 612.5 & 84.5 & 62.2 & 53.0 & 96.0 & 76.1 & 87.4 & 79.0 & 80.8 & 75.4 & 77.2 \\
        \midrule
        % \rowcolor{gray!15}
        \multicolumn{13}{l}{\textbf{\textit{Ours}}} \\ 
        \modelbase & Vicuna-7B & 0.6 &  110.8 & 79.6 & 63.0 & 52.9 & 67.8 & 58.4 & 86.9 & 75.2 & 64.8 & 67.5 & 68.5 \\
        \multicolumn{2}{c|}{ \textcolor{black!60}{$\Delta$ \textit{compare to} LLaVA-v1.5}} & \textcolor{ForestGreen}{\textbf{7.1\%}} & \textcolor{ForestGreen} {\textbf{43.6\%}}  & \textcolor{ForestGreen} {\textbf{+1.1}} & \textcolor{ForestGreen} {\textbf{+1.0}} & \textcolor{ForestGreen} {\textbf{+2.9}} & \textcolor{ForestGreen} {\textbf{+1.0}} & \textcolor{ForestGreen} {\textbf{+0.2}} & \textcolor{ForestGreen} {\textbf{+1.0}} & \textcolor{BrickRed} {\textbf{-0.3}} & \textcolor{ForestGreen} {\textbf{+0.5}} & 
        \textcolor{ForestGreen} {\textbf{+1.4}} & \textcolor{ForestGreen} {\textbf{+1.0}}
        \\
        \modelhighres & Vicuna-7B & 1.7 & 148.6 & 81.4 &  63.7 & 57.8 & 71.7 & 64.3 & 87.0 & 77.5 & 68.1 & 71.6 & 71.5\\
        \multicolumn{2}{c|}{ \textcolor{black!60}{$\Delta$ \textit{compare to} LLaVA-v1.6}} & \textcolor{ForestGreen}{\textbf{5.2\%}} & \textcolor{ForestGreen}{\textbf{29.6\%}} & \textcolor{BrickRed} {\textbf{-0.4}} & \textcolor{BrickRed} {\textbf{-0.5}} & \textcolor{ForestGreen} {\textbf{+0.2}} & \textcolor{ForestGreen} {\textbf{+1.6}} & \textcolor{BrickRed} {\textbf{-0.6}} & \textcolor{ForestGreen} {\textbf{+0.5}} & \textcolor{ForestGreen} {\textbf{+1.5}} & \textcolor{ForestGreen} {\textbf{+0.7}} & \textcolor{ForestGreen} {\textbf{+1.4}} & \textcolor{ForestGreen} {\textbf{+0.5}}
        \\
        \modelvideo & Qwen2-7B & 3.6 & 198.1 & 84.0 &  65.0 & 53.8 & 95.4 & 77.0 & 87.1 & 80.9 & 79.3 & 76.9 & 77.7 \\ 
        \multicolumn{2}{c|}{ \textcolor{black!60}{$\Delta$ \textit{compare to} LLaVA-OV}} & \textcolor{ForestGreen}{\textbf{6.0\%}} & \textcolor{ForestGreen}{\textbf{32.3\%}} & \textcolor{BrickRed} {\textbf{-0.5}} & \textcolor{ForestGreen} {\textbf{+2.8}} & \textcolor{ForestGreen} {\textbf{+0.8}} & \textcolor{BrickRed} {\textbf{-0.6}} & \textcolor{ForestGreen} {\textbf{+0.9}} & \textcolor{BrickRed} {\textbf{-0.3}} & \textcolor{ForestGreen} {\textbf{+1.9}} & \textcolor{BrickRed} {\textbf{-1.5}} & \textcolor{ForestGreen} {\textbf{+1.5}} & \textcolor{ForestGreen} {\textbf{+0.5}}
        \\
        \bottomrule
    \end{tabular}
     }
    \vspace{-12pt}
    \label{tab:image_bench}
\end{table*}

\subsection{Experimental Settings}

\paragraph{Implementation Details.} 
We sequentially train three models to investigate the potential and scalability of the proposed architecture. We begin with \modelbase, which mirrors the configuration of LLaVA-v1.5 \citep{liu2023improvedllava}, using the CLIP ViT-L/336px \citep{radford2021learning} as the vision encoder and Vicuna-v1.5-7B~\citep{chiang2023vicuna} as the LLM backbone. For high-resolution scalability, we incorporate a dynamic high-resolution mechanism adopted in LLaVA-v1.6~\citep{liu2024llavanext} for fair comparison, resulting in \modelhighres. Furthermore, to explore the full potential of the proposed approach, we extend it to an advanced version, \modelvideo, which is capable of handling both image and video understanding. For \modelvideo, in line with LLaVA-OneVision \citep{li2024llava}, we replace the vision encoder with the SigLIP ViT-SO400M/384px \citep{zhai2023sigmoid} and use Qwen2-7B-Instruct \citep{yang2024qwen2} as the LLM backbone. For all three variants, we uniformly select 25\% of the layers in the LLMs and replace their original LN modules with ViLN, upon which FMI is applied. We adopt the attention-based conditioning as the default method.  For image inputs, the full outputs of the vision encoder are utilized for the vision conditioning. For video inputs, 32 frames are uniformly sampled. All experiments are conducted on 16 NVIDIA A100 GPUs, with the training hyperparameters detailed in the Appendix.

\paragraph{Training Data.} 
(1) Pre-training Datasets. We train all three \model variants using publicly available images from CC12M \citep{changpinyo2021conceptual}. Following the pre-processing pipeline outlined in \citep{radford2021learning}, we retain only samples with resolutions exceeding 448 × 448, resulting in a curated subset of 8M samples. (2) Supervised Fine-tuning Datasets. For \modelbase, we leverage the instruction datasets corresponding to LLaVA-v1.5 \citep{liu2023improvedllava}, \textit{i.e.}, LLaVA-665K. For \modelhighres, we leverage the instruction datasets corresponding to LLaVA-v1.6 \citep{liu2024llavanext}, \textit{i.e.}, LLaVA-760K. For \modelvideo, we leverage the instruction data from LLaVA-OneVision \citep{li2024llava}. For further details, please refer to the Appendix.

\paragraph{Evaluation Benchmarks and Metrics.}
We evaluate \model on both image and video understanding tasks, including 9 image benchmarks and 6 video benchmarks. For evaluation metrics, we report two categories: \textit{computational efficiency} and \textit{benchmark accuracy}. Specifically, computational efficiency includes FLOPs (T) and latency (ms). Further details could be found in the Appendix.

\subsection{Evaluation Results}
\paragraph{Image Understanding Evaluation.}

We compare \model with baseline models across 9 benchmarks to assess its efficiency and performance, with results presented in Table \ref{tab:image_bench}. \model strikes a remarkable balance between computational efficiency and performance when compared with all baseline models. The three \model variants—\modelbase, \modelhighres, and \modelvideo—are compared against LLaVA-v1.5, LLaVA-v1.6, and LLaVA-OV, respectively. Compared to their counterparts, they achieve reductions of 14.0×, 19.4×, and 16.8× in FLOPs cost. Despite substantial reductions in computational overhead,  the three variants
achieve 1.0\%, 0.5\%, and 0.5\% average accuracy improvements across all benchmarks, respectively.
These results underscore the superior cross-modal interaction efficiency of FMI compared to existing integration strategies. A more comprehensive comparison of the three strategies is provided in Table~\ref{tab:archi} of Section~\ref{sec:arc_compare} for further reference.

\begin{table*}[t]
    \centering
    \caption{\textbf{Performance on 6 video-based benchmarks,} including EgoSchema, MLVU, VideoMME, MVBench, CinePile and Video-ChatGPT. Along with computational efficiency and accuracy metrics, we also report the number of sampled frames for each video.} 
    \setlength{\tabcolsep}{2.5pt} % 缩小列间距
    \resizebox{\linewidth}{!}{  
    \begin{tabular}{lc| cc| cccccc}
        \toprule
        \multirow{2}{*}{\textbf{Method}} &
        \multirow{2}{*}{\textbf{\#Frames}} &
        \multicolumn{2}{c|} {\textbf{Efficiency}} & \multicolumn{6}{c} {\textbf{Performance}} \\ 
        \cmidrule(lr){3-4} \cmidrule(lr){5-10}

        & & \textbf{FLOPs} & \textbf{Latency} & \textbf{EgoSchema} & \textbf{MLVU} & \textbf{VideoMME} & \textbf{MVBench} & \textbf{CinePile} & \textbf{Video-ChatGPT}  \\
        \midrule 
        Video-LLaVA \citep{lin2023video} & 8 & 32.6 & 488.6 & 38.4 & 47.3 & 39.9 & 43.1 & 25.7 & 2.84 \\
        % VideoChat2\citep{li2024mvbench} & 16 & 28.8 & & 54.4 & 47.9 & 54.6 &  60.4 & 29.3 & 2.95  \\
        ShareGPT4Video~\citep{chen2024sharegpt4video} & 16 & 39.2 & 502.7 & -- & 46.4 & 43.6 & 51.2 & -- & -- \\
        VideoLLaMA2~\citep{damonlpsg2024videollama2} & 16 & 27.3 & 465.5 & 51.7 &  48.5 & 46.6 & 54.6 & 44.6 & --  \\
        LongVA~\citep{zhang2024long} & 32  & 84.5 & 742.2 & -- & -- & 51.8 & -- & 41.0 & 3.17 \\
        LLaVA-NeXT-Video~\citep{liu2024llavanext} & 32 & 89.6 & 775.4 & 43.9 & -- & 33.7 & 46.5 & -- & -- \\
        LLaVA-OV~\citep{li2024llava} & 32 & 129.6 & 1215.6 & 60.1 & 64.7 & 58.2 & -- & 49.3 & 3.49 \\
        % Video-ChatGPT~\citep{maaz2023video} & 100 & 21.8 & & 36.2 & 31.3 & -- & 32.7 & 15.1 & 2.37 \\
        LLaMA-VID~\citep{li2024llama} & 1fps & 182.1 & 2174.3 &  38.5 & 33.2 & 25.9 & 41.9 & -- & 2.88 \\
        \midrule
        \modelvideo (Ours) & 8 & 4.2 & 217.0 & 51.8 & 54.2 & 49.4 & 51.8 & 45.6 & 3.03 \\
        \modelvideo (Ours) & 16 & 8.9 & 272.3 & 55.5 & 58.5 & 54.0 & 54.3 & 50.3 & 3.14 \\
        \modelvideo (Ours) & 32 & 18.6 & 401.5 & 58.4 & 62.3 & 57.3 & 56.5 & 54.0 & 3.23  \\
        \bottomrule
    \end{tabular}
     }
    \vspace{-12pt}
    \label{tab:video_bench}
\end{table*}
\paragraph{Video Understanding Evaluation.}
We compare \modelvideo with advanced video baseline models across 6 widely used benchmarks. To conduct a more comprehensive comparative analysis, in addition to the default setting of 32 frames, we further train two versions utilizing 8 and 16 frames, respectively. The results are presented in Table \ref{tab:video_bench}. The superiority of \modelvideo is strikingly clear. It demonstrates significant computational efficiency, achieving a 6× to 7× reduction in FLOPs compared to baseline models with identical frame counts. Notably, the FLOPs required for the 32-frame \modelvideo are comparable to half of those needed by the 8-frame Video-LLaVA~\citep{lin2023video}. \modelvideo also consistently surpasses or matches the baseline models in performance across all tested frame configurations. We further provide a thorough analysis of the computational overhead associated with frame extension in Section \ref{sec:frame_extend}.

\subsection{Ablation Study}

In this section, we conduct a comprehensive ablation study of the proposed method. For all experiments in this section, we adopt SigLIP ViT-SO400M~\citep{zhai2023sigmoid} as vision encoder and Qwen2-7B-Instruct~\citep{yang2024qwen2} as LLM backbone. For training data, we uniformly leverage a 4M subset of the pretraining dataset and LLaVA-665K for alignment and SFT, respectively.

\paragraph{Comprehensive Comparison of Integration Techniques.}
\label{sec:arc_compare}
Under identical data and backbone settings, we present a comprehensive comparison of different vision information injection paradigms discussed in Figure \ref{fig:pre}. For architectural injection, we evaluate two types of inserted modules: cross-attention from Flamingo~\citep{awadalla2023openflamingo} and hyper-attention from mPLUG-Owl3\citep{ye2024mplug}.
For in-context injection, we follow the LLaVA series by concatenating visual features, mapped through a connector, into the text sequence as context. For the proposed feature modulation injection, we evaluate the three instantiations of the module introduced in Section \ref{sec:model}. The evaluation of these methods covers multiple dimensions, emphasizing computational efficiency and performance on both language-only and vision-language benchmarks. The results in Table \ref{tab:archi} demonstrate that FMI achieves a superior balance between efficiency and performance. Specifically, FMI surpasses existing paradigms in both training time and inference overhead. Additionally, we plot the learning curves of the three injection paradigms during the pre-training phase, as illustrated in Figure \ref{fig:loss}. FMI achieves faster convergence. With the same pre-training dataset, it requires only $1/8$ of the training time of in-context injection to achieve comparable performance. Furthermore, compared to current paradigms with lengthy visual contexts or additional inserted layers, FMI
preserves better linguistic proficiency,  demonstrating significant advantages on all three language-only benchmarks.

\begin{wraptable}{r}{7cm}
\centering
% \vspace{-12pt}
\caption{\textbf{Effect of modulating different sublayers.} Injecting visual information into both sublayers yields optimal results.}
    \small
    \resizebox{\linewidth}{!}{ 
    \begin{tabular}{cc|cccc|c}
        \toprule
             \textbf{Attn} & \textbf{FFN} & \textbf{VQA$^\text{T}$} & \textbf{GQA} & \textbf{MMB} & \textbf{SEED$^\text{I}$} & \textbf{Avg.} \\
        \midrule
        \ding{55} & \ding{51}  & 55.4 & 61.5 & 71.4 & 69.2 & 64.4 \\
        \ding{51}  & \ding{55}  & 57.6 & 62.4 & 72.0 & 67.8 & 65.0 \\
        \rowcolor{lightblue}  \ding{51} & \ding{51} & \textbf{58.7} & \textbf{63.2} & \textbf{72.7} & \textbf{69.5} & \textbf{66.0}\\
        \bottomrule
        \end{tabular}
        \label{tab:sublayer}
        }
    \vspace{-12pt}
\end{wraptable}

\paragraph{Effect of Modulation at Different Sublayers.}

Each LLM layer comprises two sublayers: self-attention and feed-forward. We first investigate the impact of applying ViLN at the sublayer level. Results are detailed in Table \ref{tab:sublayer}.  Disabling ViLN from either sublayer results in a performance decrement, notably more pronounced when removed from the self-attention sublayer. This observation likely stems from the self-attention sublayer's pivotal role in handling interactions between tokens, having a more substantial influence on the efficacy of cross-modal interactions.

\begin{table*}[t]
    \centering
    \small
    \begin{center}
    \caption{\textbf{Comparison of integration techniques under identical data and backbone settings.} We present the total training hours (Time), the FLOPs during inference, and the accuracy results on three language-only benchmarks and four vision-language tasks.}
    \setlength{\tabcolsep}{3pt} % 缩小列间距
    \resizebox{\linewidth}{!}{  
    \begin{tabular}{lccccccccccc}
        \toprule
        \multirow{2}{*}{\textbf{Architecture}} & \multicolumn{2}{c}{\textbf{Efficiency}} & \multicolumn{4}{c}{\textbf{Language Benchmarks}} & \multicolumn{5}{c}{\textbf{Vision-Language Benchmarks}} \\ 
        
        \cmidrule(lr){2-3} \cmidrule(lr){4-7} \cmidrule(lr){8-12}
        
         & \textbf{Training Time} & \textbf{FLOPs} & \textbf{MMLU} & \textbf{MBPP} & \textbf{MATH} &  \textbf{Avg.} & \textbf{VQA$^\text{T}$} & \textbf{GQA} & \textbf{MMB} & \textbf{SEED$^\text{I}$} &  \textbf{Avg.} \\ 
        \midrule

         \color{gray}{Qwen2-7B-Instruct} &  \color{gray}{--} &  \color{gray}{--} & \color{gray}{69.3} &  \color{gray}{66.2} &  \color{gray}{47.8} &  \color{gray}{61.1} & \color{gray}{--} & \color{gray}{--} & \color{gray}{--} &  \color{gray}{--} & \color{gray}{--} \\
        \midrule
        \multicolumn{12}{l}{\textbf{\textit{Architectural Injection}}} \\
        \quad Cross Attention  & 9.8 & 2.5  & 64.8 & 62.4 & 40.8 & 56.0 &  55.8 &  62.4 & 71.6 & 68.0 & 64.5\\ 
        \quad Hyper Attention  & 8.7 & 2.3  & 65.3 & 62.0 & 41.2 & 56.2 & 56.6 & 61.8 & 71.8 & 68.4 & 64.7 \\

        \multicolumn{12}{l}{\textbf{\textit{In-context Injection}}} \\
        \quad Concat  & 22.0 & 11.4  & 66.2 & 63.6 & 42.4 & 57.4 & \textbf{59.0} & \textbf{63.4} & 72.0 & 69.2 & 65.9 \\

        \multicolumn{12}{l}{\textbf{\textit{Feature Modulation Injection}}} \\
    
         \quad MLP-based & \textbf{5.8}  & \textbf{0.8} & 
 \textbf{68.4} & \textbf{66.0} & \textbf{45.2} & \textbf{59.9} & 58.4 & 63.0 & 72.1 & 68.6 & 65.5 \\
        \quad Conv-based  & 6.0 & 0.8 & 67.7 & 65.4 & 44.9 & 59.3 & 58.0 & 62.7 & 72.4 & 67.5 & 65.2 \\

        \rowcolor{lightblue} \quad Attention-based  & 6.6  & 0.9  & 
 68.2 & 65.6 & 44.6 & 59.5 & 58.7 & 63.2 & \textbf{72.7} & \textbf{69.5} &\textbf{66.0}\\
        \bottomrule
    \end{tabular}
    \label{tab:archi}
    }
    \end{center}
    \vspace{-12pt}
\end{table*}

\begin{wraptable}{r}{7cm}
\vspace{-12pt}
\centering
\caption{\textbf{Effect of modulation parameters}. Each parameter enhances visual information integration through corresponding operation.}
    \small
    \resizebox{\linewidth}{!}{ 
    \begin{tabular}{cc|cccc|c}
        \toprule
              \textbf{$\Delta \alpha_{\bm{v}}$} & \textbf{$\Delta \beta_{\bm{v}}$} &\textbf{VQA$^\text{T}$} & \textbf{GQA} & \textbf{MMB} & \textbf{SEED$^\text{I}$} & \textbf{Avg.} \\
        \midrule
        \ding{51} & \ding{55} & 58.1 & 62.2 & 70.8 & 67.7 & 64.7\\
         \ding{55} & \ding{51} & \textbf{59.3} & 62.7 & 69.3 & 66.5 & 64.5 \\
        \rowcolor{lightblue} 
         \ding{51} & \ding{51} & 58.7 & \textbf{63.2} & \textbf{72.7} & \textbf{69.5} & \textbf{66.0} \\ 
        \bottomrule
        \end{tabular}
        \label{tab:parameter}
    }
    \vspace{-12pt}
\end{wraptable}

\paragraph{Effect of Modulation Parameter.}
ViLN involves two vision-conditioned affine parameters $\Delta \alpha_{\bm{v}}$ and $\Delta \beta_{\bm{v}}$, which adaptively perform scaling and shifting operations on the internal hidden states. We present an ablation study of each affine parameter, with findings detailed in Table \ref{tab:parameter}. The results indicate that each parameter positively impacts the overall performance, and the gains attributed to each parameter are comparable. It demonstrates that both addictive and multiplicative operations hold similar importance for the effective integration of visual information.

\begin{wraptable}{r}{7cm}
\vspace{-12pt}
\centering
\caption{\textbf{Effect of modulation frequency}. Applying ViLN at a moderate frequency yields the best average performance.}
    \small
    \resizebox{\linewidth}{!}{ 
    \begin{tabular}{c|cccc|c}
        \toprule
        \textbf{Frequency} & \textbf{VQA$^\text{T}$} & \textbf{GQA} & \textbf{MMB} & \textbf{SEED$^\text{I}$} & \textbf{Avg.} \\
        \midrule
        100\% & \textbf{59.1} & 62.9 & 71.9 & 68.2 & 65.5 \\
        50\%  & 58.1 & 62.5 & 70.9 & 69.1 & 65.2  \\
        \rowcolor{lightblue} 25\%   & 58.7 & \textbf{63.2} & \textbf{72.7} & \textbf{69.5} & \textbf{66.0} \\
        12.5\% & 57.6 & 62.2 & 71.5 & 67.0 & 64.6 \\
        \bottomrule
    \end{tabular}
        \label{tab:ratio}
        }
    \vspace{-12pt}
\end{wraptable}

\paragraph{Effect of Modulation Frequency.}
We investigate the effect of modulation frequency, defined as the proportion of layers within the LLM that apply ViLN. Specifically, we vary this frequency from 12.5\% to 100\% and report the corresponding results in Table~\ref{tab:ratio}. Notably, applying ViLN to 25\% of the layers achieves the best average performance across benchmarks, which suggests that a moderate frequency is sufficient to effectively inject visual context, while higher frequencies may not lead to further gains.

\begin{wraptable}{r}{7cm}
\vspace{-12pt}
\centering
\caption{\textbf{Effect of modulation location}. Uniformly distributing ViLN across layers leads to more effective multimodal integration.}
    \small
    \resizebox{\linewidth}{!}{ 
    \begin{tabular}{c|cccc|c}
        \toprule
        \textbf{Location} & \textbf{VQA$^\text{T}$} & \textbf{GQA} & \textbf{MMB} & \textbf{SEED$^\text{I}$} & \textbf{Avg.} \\
        \midrule
        shallow & 54.7 & 59.4 & 69.2 & 64.5 & 62.0 \\
        middle  & 56.5 &  61.6 & 71.4 & 67.3 & 64.2\\
        deep   & 57.0 & 60.8 & 70.1 & 65.9 & 63.4 \\
        \rowcolor{lightblue} uniform & \textbf{58.7} & \textbf{63.2} & \textbf{72.7} & \textbf{69.5} & \textbf{66.0}  \\
        \bottomrule
    \end{tabular}
        \label{tab:location}
        }
    % \vspace{-12pt}
\end{wraptable}

\paragraph{Effect of Modulation Location.}
We further study the effect of modulation location by fixing the modulation frequency at 25\% and varying the specific layers within the LLM where ViLN is applied. Four strategies are evaluated: shallow (first 25\%), deep (last 25\%), middle (central 25\%), and uniform (evenly spaced). As shown in Table~\ref{tab:location}, the uniform strategy consistently outperforms the others, indicating that distributing  modulation across different depths enables more effective and balanced vision integration.

\subsection{Visualization and Analysis}

\begin{figure*}[t]
    \centering
    \begin{minipage}{0.32\textwidth}
        \centering
    
  \includegraphics[width=\linewidth]{ 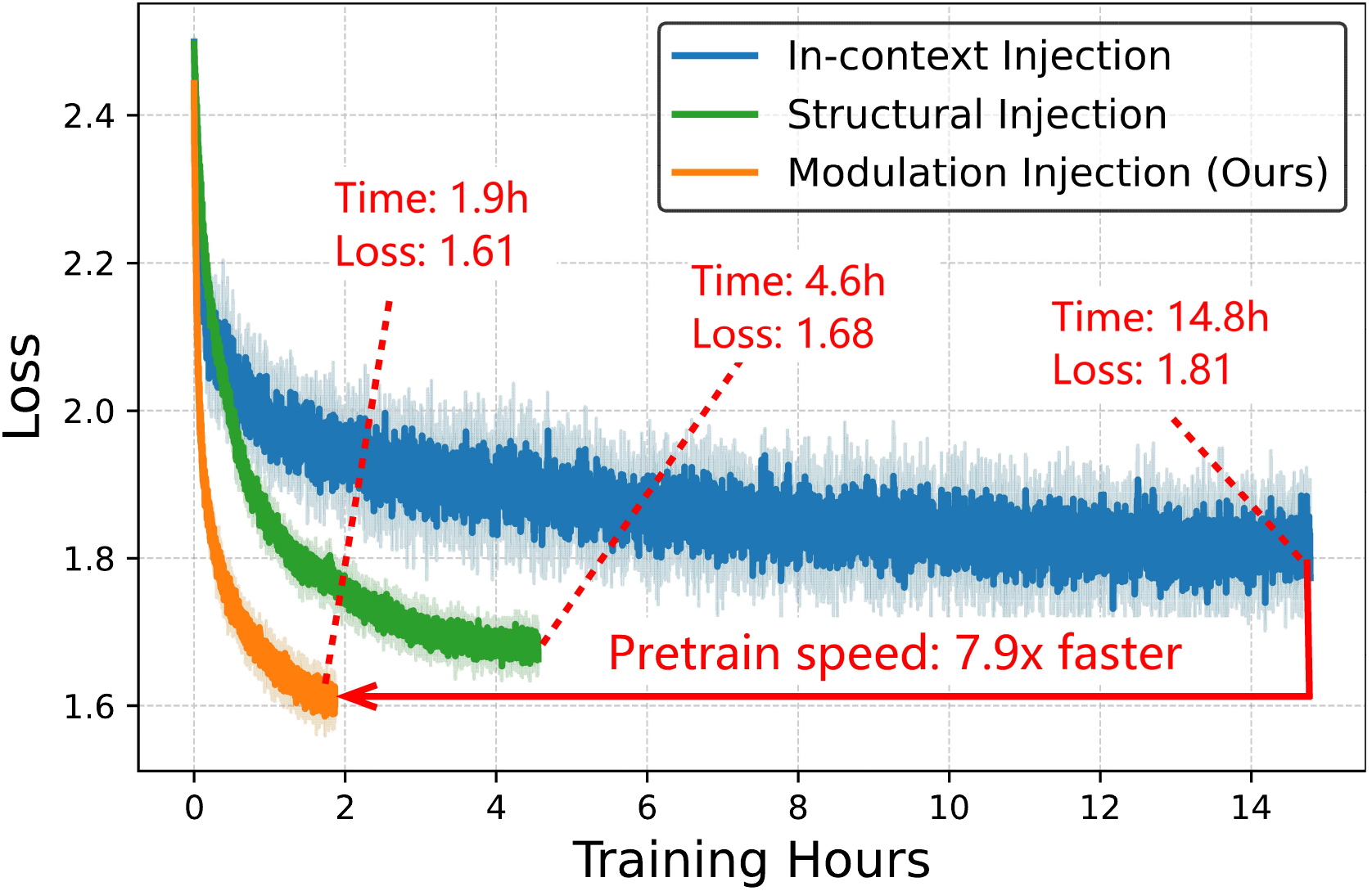}
        \caption{\textbf{Training loss of three injection techniques over time.} Our method achieves faster convergence and lower loss.}
        \label{fig:loss}
    \end{minipage}
    \hfill
    \begin{minipage}{0.32\textwidth}
        \centering
        \includegraphics[width=\linewidth]{ 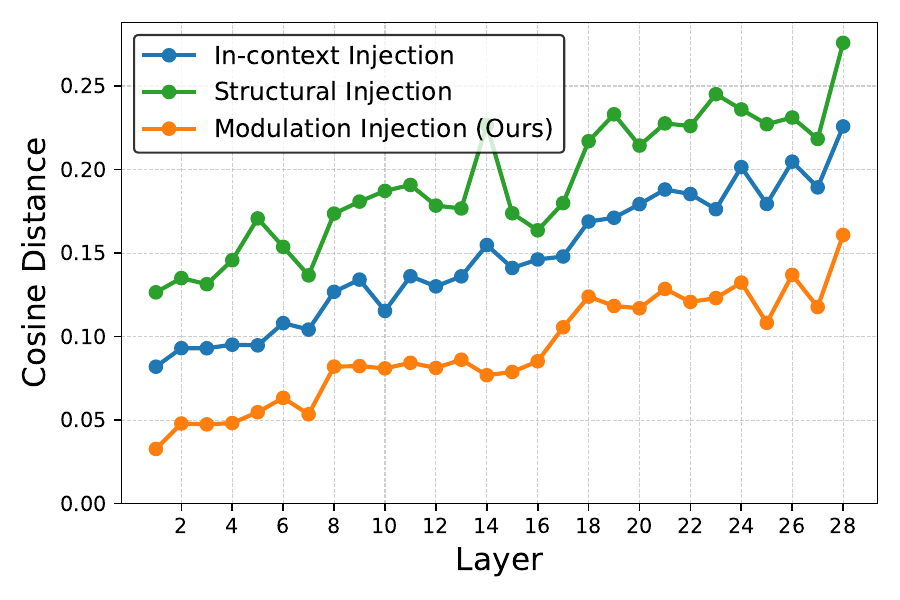}
        \caption{\textbf{Feature drifts compared with base LLM.} Our method preserves best linguistic capabilities.}
        \label{fig:text}
    \end{minipage}
    \hfill
    \begin{minipage}{0.32\textwidth}
        \centering
        \includegraphics[width=\linewidth]{ 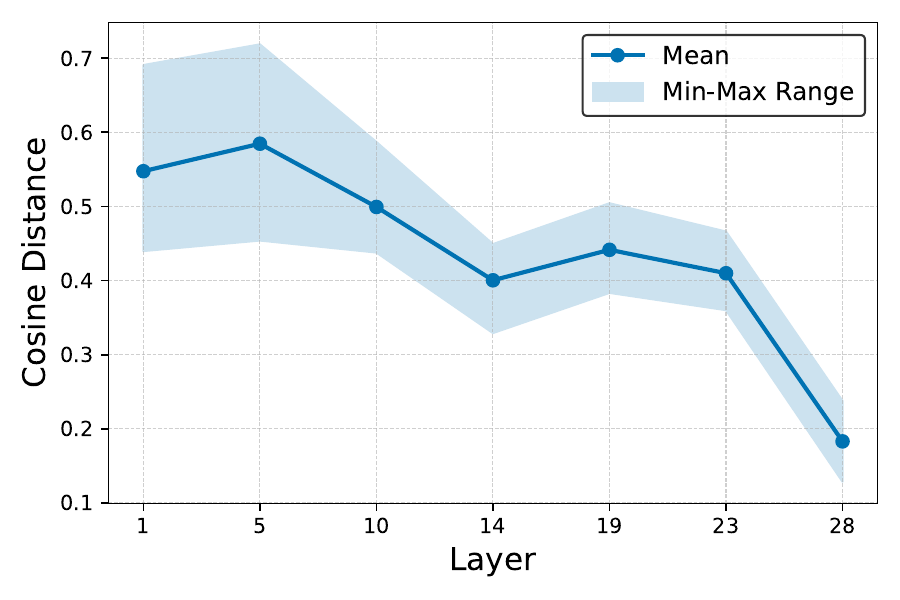}
        \caption{\textbf{Features distances before and after ViLN across layers.} Stronger in early layers, while stabilize in deeper layers.}
        \label{fig:layer}
    \end{minipage}
\end{figure*}

\paragraph{Effective Preservation of Linguistic Capabilities.} 
For the baselines compared in Table~\ref{tab:archi}, we measure the cosine distance between their hidden states and those of the original base LLM (\textit{i.e.}, Qwen2-7B-Instruct~\citep{yang2024qwen2}) on QA samples from a language-only benchmark, MMLU~\citep{hendrycks2020measuring}.  We plot the layer-wise distances in Figure~\ref{fig:text}.  \model exhibits minimal feature drift compared to the base LLM, which directly correlates with performance on language-only benchmarks as reported in Table~\ref{tab:archi}. It visually demonstrates that FMI effectively preserves linguistic capabilities.

\paragraph{Stronger Modulation Influence  on Early Layers.} 
For \model, we compute the cosine distance between features before and after modulation at each layer as a metric for modulation influence on GQA~\citep{hudson2019gqa}. Figure \ref{fig:layer} shows the average distance (solid line) and the range (shaded area) across tokens. The tokens in early layers undergo significant modulation with notable variance among tokens, while deeper layers show reduced influence and stability. This reflects early layers dynamically establishing cross-modal alignments, while deeper layers refine them into coherent representations.

\paragraph{Stronger Modulation Influence on Semantically Rich Tokens.} 

\begin{figure*}[t]
    \centering
    \begin{minipage}{0.32\textwidth}
        \centering
        \includegraphics[width=\linewidth]{ 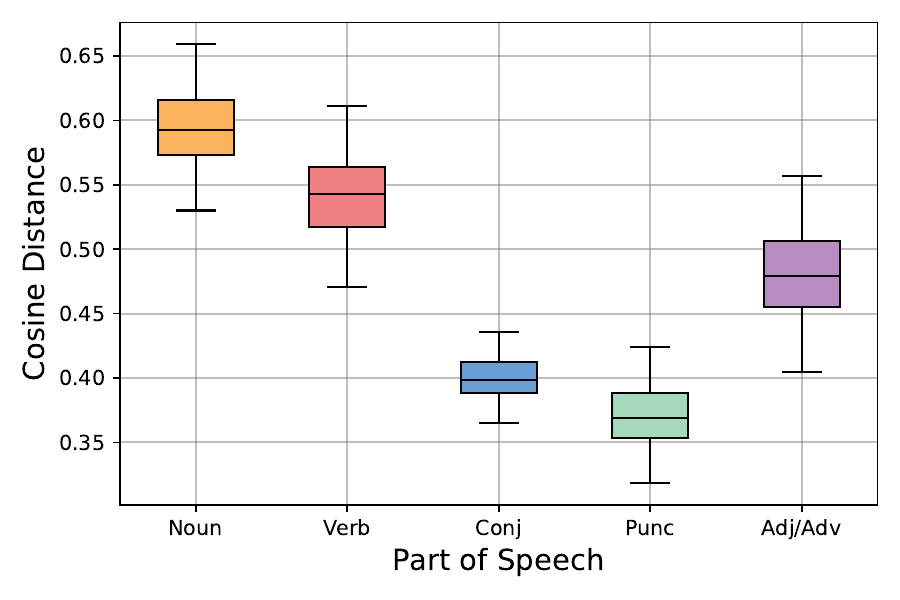}
        \caption{\textbf{Modulation influence of \model across POS categories.} Semantically rich tokens exhibit stronger modulation influence.}
        \label{fig:word}
    \end{minipage}
    \hfill
    \begin{minipage}{0.32\textwidth}
        \centering
        \includegraphics[width=\linewidth]{ 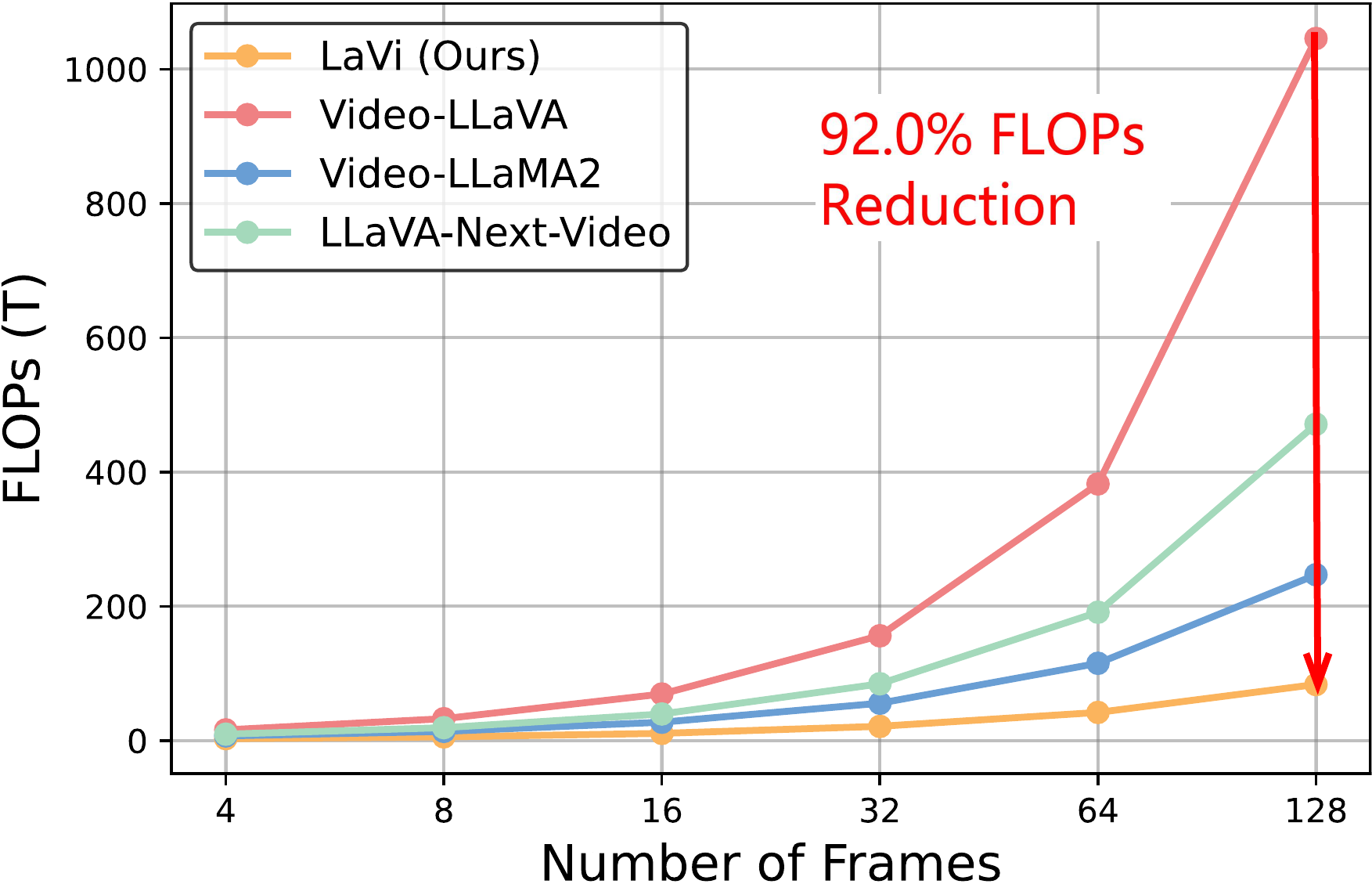}
    \caption{\textbf{FLOPs comparison across frame counts.} \modelvideo achieves significant FLOPs reduction across all frames.}
        \label{fig:flop}
    \end{minipage}
    \hfill
    \begin{minipage}{0.32\textwidth}
        \centering
        \includegraphics[width=\linewidth]{ 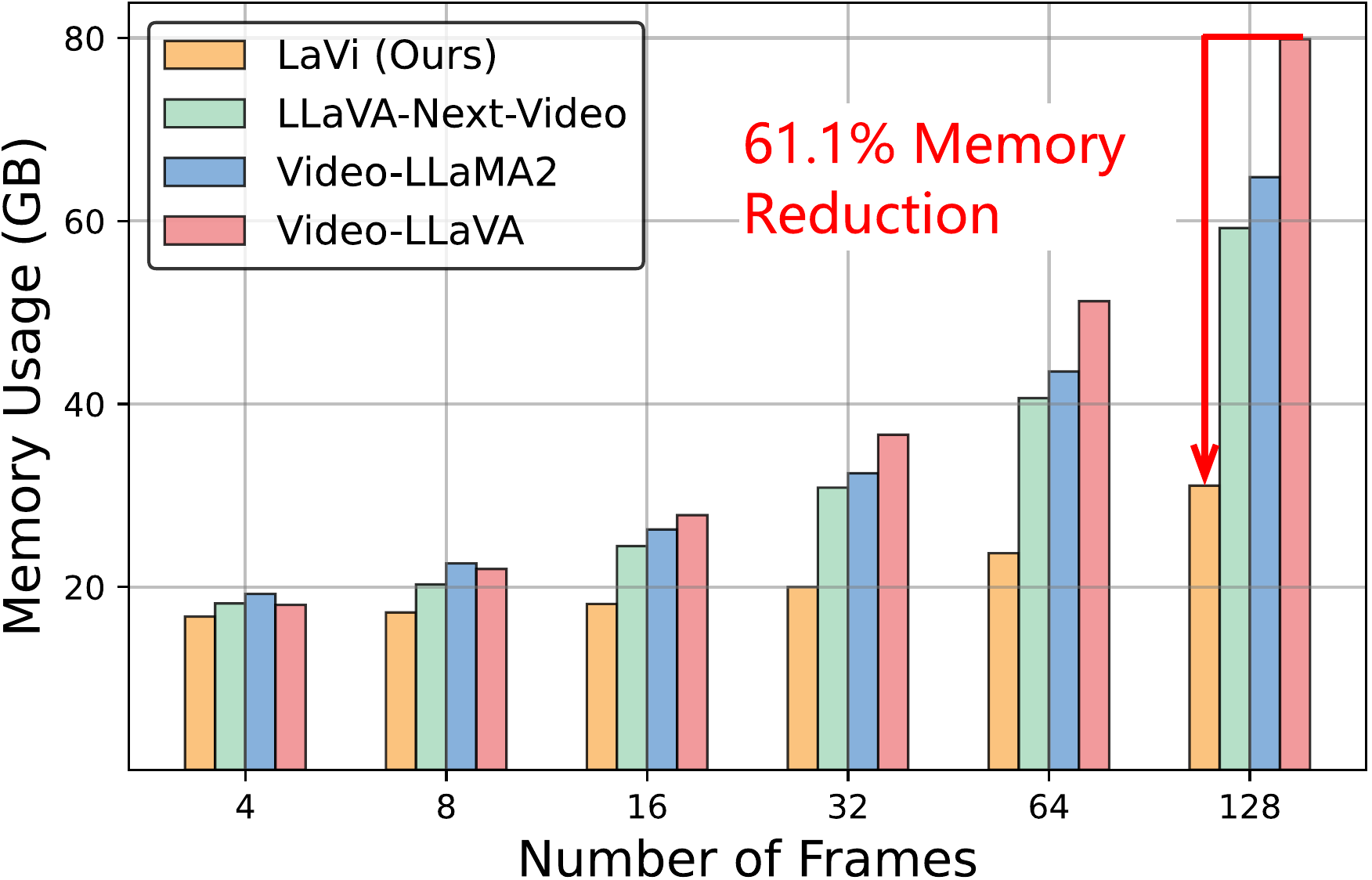}
      \caption{\textbf{Memory comparison across frame counts.} \modelvideo achieves significant Memory reduction across all frames.}
        \label{fig:vram}
    \end{minipage}
\end{figure*}

In addition to layer-wise influence, we evaluate the cosine distances across different token types, specifically part-of-speech (POS) categories: nouns, verbs, conjunctions, adjectives/adverbs, and punctuation. POS tagging is performed using the NLTK toolkit~\citep{bird2006nltk}. As illustrated in Figure \ref{fig:word}, nouns and verbs exhibit more significant modulation influence compared to conjunctions and punctuation. This is intuitive, as nouns and verbs, which carry richer semantic meaning, are more likely to align with and integrate visual information during cross-modal interactions.

\paragraph{Superior Vision Sequence Scalability.} 
\label{sec:frame_extend}
High-resolution images and videos increase visual sequence lengths, leading to higher computation and memory demands. We compare FLOPs and GPU memory usage between \modelvideo and baseline models~\citep{damonlpsg2024videollama2,lin2023video,liu2024llavanext} as frame counts grow, shown in Figure \ref{fig:flop} and \ref{fig:vram}.  \modelvideo achieves significant computational overhead reduction, saving 92.0\% and 61.1\% in FLOPs and memory, respectively, at 128 frames compared to Video-LLaVA, while maintaining superior performance on video understanding benchmarks. These results demonstrate \modelvideo's exceptional computational scalability over existing paradigms.

% \begin{figure}[t]
%     \centering
%     \begin{minipage}{\linewidth}
%         \centering      \includegraphics[width=\linewidth]{ICCV2025-Author-Kit/figures/layer_plot.pdf}
%         \vspace{-2.5mm}
%     \end{minipage}%
%      \vspace{-3mm}
%     \caption{\textbf{Layer comparison.} The extent of affine transformation effects across different LLM layers.}
%     \label{fig:layer}
% \end{figure}

\section{Conclusion}
In this work, we propose a novel internal feature modulation injection paradigm for LVLMs, ensuring minimal structural interference and superior computational scalability by avoiding excessive context expansion. Building on this paradigm, we develop \model, a highly efficient LVLM that leverages Vision-Infused Layer Normalization (ViLN) for precise visual-linguistic alignment while drastically reducing computational costs. Compared to LLaVA-style models, \model achieves 94.0\% FLOP reduction, runs 3.1× faster, and significantly lowers latency, establishing LaVi as a highly efficient alternative for vision-language integration.

% In this work, we present \model, a highly efficient and minimalistic architecture for LVLMs that addresses the trade-off between performance and computational overhead. By conditioning layer normalization on visual features, we enable seamless and lightweight cross-modal interaction without modifying the core structure of the language model. Our extensive experiments across multiple image and video understanding benchmarks demonstrate that \model achieves state-of-the-art performance while significantly reducing computational complexity. With a 15× reduction in FLOPs compared to prior models, \model sets a new standard for efficient, low-latency multimodal processing. This approach not only advances the practical deployment of LVLMs but also opens avenues for further research in optimizing multimodal models for real-time applications.

% \section*{References}
\bibliography{neurips_2025}
\bibliographystyle{plain}

%%%%%%%%%%%%%%%%%%%%%%%%%%%%%%%%%%%%%%%%%%%%%%%%%%%%%%%%%%%%
% \clearpage
\clearpage
\setcounter{page}{1}
% \maketitlesupplementary
\begin{appendix}

\section{Implementation Details}
% \begin{table}[htbp]
%     \centering\small
%     \caption{
%     \textbf{Training details of \model.}}
%     \label{tab:impl}
%     \setlength{\tabcolsep}{5pt}
%     \resizebox{0.7\linewidth}{!}{  
%     \begin{tabular}{l cc}
%     \toprule
%     Config & Stage I & Stage II \\
%     \midrule
%     Trainable parts & Condition Module & Condition Module  + LLM \\
%     Frozen parts & ViT + LLM & NA \\
%     Global batch size & 1024 & 256 \\
%     Batch size per GPU & 64 & 16 \\
%     Accumulated steps & 1 & 1 \\
%     DeepSpeed zero stage & 2 & 2 \\
%     Learning rate & 1$\times$10$^{-\text{3}}$ & 2$\times$10$^{-\text{5}}$ \\
%     Learning rate schedule & \multicolumn{2}{c}{warmup + cosine decay} \\
%     Warmup ratio & \multicolumn{2}{c}{0.03} \\
%     Weight decay & \multicolumn{2}{c}{0} \\
%     Epoch & \multicolumn{2}{c}{1} \\
%     Optimizer & \multicolumn{2}{c}{AdamW} \\
%     Precision & \multicolumn{2}{c}{bf16} \\
%     \bottomrule
%     \end{tabular}
%     }
% \end{table}

\subsection{Trianing Details.}

The overall training process adopts a two-stage paradigm, initially involving the pretraining of the conditioning module, followed by instruction tuning. Table \ref{tab:impl1} and Table \ref{tab:impl2} presents the details of this two-stage training for \model. The implementation includes two sets of vision and LLM combinations: CLIP ViT-L/336px~\citep{radford2021learning} + Vicuna~\citep{chiang2023vicuna} or SigLIP ViT-SO400M/384px~\citep{zhai2023sigmoid} + Qwen2~\citep{yang2024qwen2}, aligned with the respective LLaVA configurations. Furthermore, consistent with the settings of LLaVA1.6 and LLaVA-OV, we additionally unfroze the ViT during the SFT phase.

\subsection{Benchmark Details.}

We conduct a comprehensive evaluation of \model, including both image and video understanding benchmarks.

\paragraph{Image-based Benchmarks}

Following the LLaVA framework~\citep{liu2023improvedllava}, we conduct experiments across 9 widely recognized benchmarks, including VQA-v2 ($\text{VQA}^{\text{v2}}$)~\citep{goyal2017making}, GQA~\citep{hudson2019gqa}, VisWiz~\citep{gurari2018vizwiz}, ScienceQA-IMG (SciQA) \citep{lu2022learn}, TextVQA ($\text{VQA}^{\text{T}}$)~\citep{singh2019towards}, POPE~\citep{li2023evaluating}, MME~\citep{fu2023mme}, MMBench (MMB) \citep{liu2024mmbench}, SEED-Bench ($\text{SEED}^{\text{I}}$) \citep{li2024seed}. These benchmarks span a broad spectrum of visual tasks. Our evaluation protocols are aligned with those established in the LLaVA framework, ensuring fair consistency.

\paragraph{Video-based Benchmarks}

We conduct experiments across 6 widely recognized benchmarks, including MVBench~\citep{li2024mvbench}, MLVU~\citep{zhou2024mlvu}, EgoSchema~\citep{mangalam2023egoschema}, VideoMME~\citep{fu2024video}, CinePile~\citep{rawal2024cinepile} and Video-ChatGPT~\citep{maaz2023video}. They cover multiple knowledge dimensions and domain focuses, with video durations ranging from a few seconds to several hours. 

\subsection{Evaluation Details.}

We adopt LMMs-Eval as our evaluation toolkit. For evaluation prompts, we provide a thorough examination of all evaluation benchmarks utilized in this paper in Table~\ref{tab:prompt}. For model efficiency, the FLOPs and latency are calculated using the DeepSpeed toolkit \citep{deepspeed_fps} on a single A100 GPU without any engineering acceleration techniques.

\begin{table}[htbp]
  \centering
  %----------------------------- 左侧表 -----------------------------%
  \begin{minipage}[t]{0.45\linewidth}
    \centering
    \caption{The training details of \model based on Vicuna.}
    \label{tab:impl1}
    \setlength{\tabcolsep}{5pt}
    \resizebox{\linewidth}{!}{%
      \begin{tabular}{l cc}
        \toprule
        Config & Stage I & Stage II \\
        \midrule
        LLM backbone       & \multicolumn{2}{c}{Vicuna-7B} \\
        ViT backbone       & \multicolumn{2}{c}{CLIP ViT-L/336px} \\
        Global batch size     & 1024                   & 256 \\
        Batch size per GPU    & 64                     & 16 \\
        Accumulated steps     & 1                      & 1 \\
        DeepSpeed zero stage  & 2                      & 2 \\
        Learning rate         & 1$\times$10$^{-3}$     & 2$\times$10$^{-5}$ \\
        Learning rate schedule& \multicolumn{2}{c}{cosine decay} \\
        Warmup ratio          & \multicolumn{2}{c}{0.03} \\
        Weight decay          & \multicolumn{2}{c}{0} \\
        Epoch                 & \multicolumn{2}{c}{1} \\
        Optimizer             & \multicolumn{2}{c}{AdamW} \\
        Precision             & \multicolumn{2}{c}{bf16} \\
        \bottomrule
      \end{tabular}}
  \end{minipage}
  \hfill
  %----------------------------- 右侧表 -----------------------------%
  \begin{minipage}[t]{0.45\linewidth}
    \centering
    \caption{The training details of \model base on Qwen.}
    \label{tab:impl2}
    \setlength{\tabcolsep}{5pt}
    \resizebox{\linewidth}{!}{%
      \begin{tabular}{l cc}
        \toprule
        Config & Stage I & Stage II \\
        \midrule
        LLM backbone       & \multicolumn{2}{c}{Qwen2-7B} \\
        ViT backbone       & \multicolumn{2}{c}{SigLIP SO400M/384px} \\
        Global batch size     & 1024                   & 256 \\
        Batch size per GPU    & 64                     & 16 \\
        Accumulated steps     & 1                      & 1 \\
        DeepSpeed zero stage  & 2                      & 3 \\
        Learning rate         & 1$\times$10$^{-3}$     & 1$\times$10$^{-5}$ \\
        Learning rate schedule& \multicolumn{2}{c}{cosine decay} \\
        Warmup ratio          & \multicolumn{2}{c}{0.03} \\
        Weight decay          & \multicolumn{2}{c}{0} \\
        Epoch                 & \multicolumn{2}{c}{1} \\
        Optimizer             & \multicolumn{2}{c}{AdamW} \\
        Precision             & \multicolumn{2}{c}{bf16} \\
        \bottomrule
      \end{tabular}}
  \end{minipage}
  \hfill
\end{table}

\begin{table}[htbp]
    \centering\small
    \caption{
    \textbf{Summary of the evaluation benchmarks.}
    Prompts are mostly borrowed from LMMs-Eval~\citep{lmms_eval2024}.
    }
    \label{tab:prompt}
    \setlength{\tabcolsep}{10pt}
    \resizebox{\linewidth}{!}{  
    \begin{tabular}{l|l}
    \toprule
    Benchmark & Response formatting prompts \\
    \midrule
    POPE \citep{li2023evaluating} & -- \\
    GQA \citep{hudson2019gqa} & Answer the question using a single word or phrase. \\
    $\text{VQA}^{\text{v2}}$ \citep{goyal2017making} & Answer the question using a single word or phrase. \\
     TextVQA \citep{singh2019towards} & Answer the question using a single word or phrase. \\
     MME \citep{fu2023mme} & Answer the question using a single word or phrase. \\
     \multirow{2}{*}{VisWiz~\citep{gurari2018vizwiz}} & 
     Answer the question using a single word or phrase. When the  \\
     & provided information is insufficient, respond with Unanswerable'.  \\
     SciQA \citep{lu2022learn} & Answer with the option's letter from the given choices directly. \\
    MMBench \citep{liu2024mmbench} & Answer with the option's letter from the given choices directly. \\
    SEED-Bench \citep{li2024seed} & Answer with the option's letter from the given choices directly. \\
    \midrule
    MLVU \citep{zhou2024mlvu}& -- \\ 
    Video-ChatGPT \citep{maaz2023video} & -- \\
    MVBench \citep{li2024mvbench}& Only give the best option. \\ 
    VideoMME \citep{fu2024video} & Answer with the option's letter from the given choices directly. \\
    EgoSchema \citep{mangalam2023egoschema} & Answer with the option's letter from the given choices directly. \\
    Cineplie \citep{rawal2024cinepile} & Answer with the option key (A, B, C, D, E) and nothing else. \\
    
    \bottomrule
    \end{tabular}}
\end{table}

\section{Conditioning Module}

To provide a clearer understanding of the proposed conditioning modules, we present PyTorch-style pseudocode implementations for the three vision-conditioned modulation strategies introduced in Section~\ref{sec:model}. Each variant—MLP-based, Conv-based, and Attention-based—is designed to instantiate the generic conditioning function $\text{Cond}(\cdot)$ used to derive token-wise affine parameters for Vision-Infused Layer Normalization (ViLN). These modules differ in how they aggregate visual context to influence individual text tokens, yet they all share a common design objective: enabling efficient, token-specific vision-language interaction without altering the LLM’s original architecture.

\begin{figure}[htbp]
\centering
{\footnotesize
\begin{lstlisting}[style=mypython]
class MLP_Condition(nn.Module):
    def __init__(self, embed_dim: int, num_vis_tok: int,
                 token_exp: int = 4, channel_exp: int = 4):
        super().__init__()
        self.L = num_vis_tok + 1  # total tokens  (t_i + v)
        # token-mixing MLP
        self.mlp_token = nn.Sequential(
            nn.Linear(self.L, self.L * token_exp),
            nn.GELU(),
            nn.Linear(self.L * token_exp, self.L))
        # channel-mixing MLP
        self.mlp_channel = nn.Sequential(
            nn.Linear(embed_dim, embed_dim * channel_exp),
            nn.GELU(),
            nn.Linear(embed_dim * channel_exp, embed_dim))
    def forward(self, t_i: torch.Tensor, v: torch.Tensor) -> torch.Tensor:
        assert v.size(1) + 1 == self.L, "Unexpected #visual tokens"
        # concat (B, L, C)   where L = 1 + V
        seq = torch.cat([t_i, v], dim=1)      # (B, L, C)
        # Token mixing
        x = seq.transpose(1, 2)  # (B, C, L)  swap token/chan
        x = self.mlp_token(x)                 
        x = x.transpose(1, 2)  # back to (B, L, C)
        # Channel mixing
        y = self.mlp_channel(y)  # (B, L, C)
        return y[:, 0, :]

class Conv_Condition(nn.Module):
    def __init__(self, embed_dim: int, kernel_size: int):
        super().__init__()
        pad = kernel_size // 2
        self.dw = nn.Conv1d(embed_dim, embed_dim, kernel_size,
                            padding=pad, groups=embed_dim)
        self.pw = nn.Conv1d(embed_dim, embed_dim, kernel_size=1)
        self.act = nn.SiLU()
    def forward(self, t_i: torch.Tensor, v: torch.Tensor) -> torch.Tensor:
        # concatenate on token dimension, then transpose for Conv1d
        seq = torch.cat([t_i, v], dim=1).transpose(1, 2)  # (B, C, 1+V)
        # depth-wise conv -> activation -> point-wise conv
        out = self.pw(self.act(self.dw(seq))).transpose(1, 2)  # (B, 1+V, C)
        return out[:, 0, :]            

class Attn_Condition(nn.Module):
    def __init__(self, C:int, h:int=8):
        super().__init__()
        self.q = nn.Linear(C, C, False)
        self.k = nn.Linear(C, C, False)
        self.v = nn.Linear(C, C, False)
        self.o = nn.Linear(C, C, False)
    def forward(self, t, v): # t:(B,1,C)  v:(B,V,C)
        B = t.size(0)
        def shp(x):
            return x.reshape(B, -1, self.h, self.dk).permute(0, 2, 1, 3)
        q, k, val = map(shp, (self.q(t), self.k(v), self.v(v)))
        attn = (q @ k.transpose(-2, -1)) / math.sqrt(self.dk)
        ctx  = (attn.softmax(-1) @ val).transpose(1, 2).reshape(B, 1, -1)
        return self.o(ctx).squeeze(1) 
\end{lstlisting}
}
	\caption{Implementation of three conditioning modules in PyTorch.}
\end{figure}

\end{appendix}

%%%%%%%%%%%%%%%%%%%%%%%%%%%%%%%%%%%%%%%%%%%%%%%%%%%%%%%%%%%%

\end{document}